# Cross-Platform Violence Detection on Social Media: A Dataset and Analysis

Celia Chen, Scotty Beland, Ingo Burghardt, Jill Byczek, William J. Conway, Eric Cotugno, Sadaf Davre, Megan Fletcher, Rajesh Kumar Gnanasekaran, Kristin Hamilton, Marilyn Harbert, Jordan Heustis, Tanaya Jha, Emily Klein[1], Hayden Kramer, Alex Leitch, Jessica Perkins, Casi Sherman, Celia Sterrn, Logan Stevens, Rebecca Zarrella, Jennifer Golbeck
University of Maryland College Park, MD
[1]Rensselaer Polytechnic Institute, Troy, NY

## ABSTRACT

Violent threats remain a significant problem across social media platforms. Useful, high-quality data facilitates research into the understanding and detection of malicious content, including violence. In this paper, we introduce a cross-platform dataset of 30,000 posts hand-coded for violent threats and subtypes of violence, including political and sexual violence. To evaluate the signal present in this dataset, we perform a machine learning analysis with an existing dataset of violent comments from YouTube. We find that, despite originating from different platforms and using different coding criteria, we achieve high classification accuracy both by training on one dataset and testing on the other, and in a merged dataset condition. These results have implications for content-classification strategies and for understanding violent content across social media.

## CCS CONCEPTS

•Information systems~World Wide Web~Web applications~Social networks •Information systems~World Wide Web~Web searching and information discovery~Content ranking

## KEYWORDS

Social media, violence, data set, machine learning

## 1 Introduction

Among the challenges of malicious online behavior, violent threats are a serious problem that can translate into offline consequences. For example, GamerGate targeted female journalists with rape and death threats over a period from 2014 to 2015 that forced some targets to cancel events and flee their homes [2]. Mass shootings and other large-scale acts of violence have arisen out of the incel community, an online extreme misogynistic movement that celebrates gender-based hate crimes [8] [12]. The January 6, 2021 U.S. insurrection drew largely on supporters of the QAnon conspiracy theory, born in online forums [14].

Moderating violent content is important ethically as well as for the culture of a platform and, potentially, for legal reasons. Training machine learning algorithms to detect this content requires high-quality datasets. There is currently one publicly available hand-coded dataset of YouTube comments annotated on the sentence level for violent threats. In this paper, we introduce the Internet Violence Study (InViS) Dataset, a new, publicly available dataset of 30,000 social media comments annotated for violence and threats of violence that covers three platforms - Reddit, Parler, and the incels.is Incel forum.

We introduce the dataset and coding criteria and then perform a cross-platform analysis with the existing YouTube data. We show that machine learning models trained on one dataset and tested on the other achieve good performance, suggesting that there is broad, cross-platform applicability of these datasets for developing violence detection algorithms.

Our research question is: Can multi-platform violence datasets be used to train models that can detect violence across platforms?

## 2 Background & Summary

The widespread presence of hate speech, harassment, and threats on social media platforms have necessitated rapid advancement in automated detection and moderation systems [9][7]. Early detection methods demonstrated success with binary classification of specific content categories like cyberbullying, outperforming broader multiclass approaches [6]  However, the sophistication of bad actors, particularly through techniques like word obfuscation, has necessitated more context-aware solutions leveraging semantic knowledge bases [1].

Machine learning approaches have shown particular promise in abusive content detection, with annotated corpora enabling increasingly sophisticated models [16]. The integration of critical race theory into annotation frameworks has improved the detection of racist and sexist content, while revealing important distinctions between hate speech and general offensive language [5] [21]. However, these advances face significant implementation challenges. The inherent subtleties of language interpretation and varying definitions of hate speech continue to complicate automated detection efforts [13]. Even well-optimized systems can increase opacity and complicate fairness considerations when deployed at scale [9].

A significant milestone in this field was the release of the YouTube Threat Corpus [10], the first publicly available dataset specifically focused on violent threats in social media comments. This dataset contains approximately 28,643 sentences from 9,845 YouTube comments collected from 19 videos covering controversial religious and political topics. Their annotation scheme, based on the WHO definition of violence, identified 1,387 sentences (4.8%) as containing violent threats or expressing sympathy with violence. This dataset has enabled benchmark evaluations for threat detection models, with Support Vector Machine classifiers achieving F1-scores of 68.85 on sentence-level classification [23].

Training data quality and availability present ongoing obstacles, with systematic reviews revealing critical gaps and biases in existing datasets [20]. These technical challenges are compounded by the social and political dimensions of content moderation, as platforms must balance free speech considerations with user protection while operating at massive scale [7]. Different approaches to this balance are evident across platforms - Reddit employs community-based moderation within platform-wide guidelines [18], while Parler and incels.is emphasize minimal intervention despite maintaining broad removal discretion [17] [11].

The human dimension of content moderation adds another layer of difficulty. Commercial content moderators face severe psychological impacts from repeated exposure to disturbing content, including PTSD symptoms, anxiety, and depression [19]. Users whose content has been moderated develop their own theories about how these systems work and actively seek ways to contest moderation decisions, revealing how content moderation shapes platform-user relationships beyond basic speech restrictions [15]. These findings suggest that while technical capabilities have advanced significantly, future developments must address not only algorithmic improvements but also the broader social and ethical implications of automated moderation systems.

## 3 Dataset

As part of this project, we have coded a new dataset for use in training and model evaluation. The InViS Dataset comprises 30,000 social media posts. These are pulled roughly equally from an incel dataset [8]  (9,936 posts), a Parler dataset [5] (10,092 posts), and the Reddit Politosphere dataset [10] (9,972).

"Violence" is a term that is used casually in a variety of contexts. We used the following definition, based on the World Health Organization's definition in the "World report on violence and health" [4].

The intentional use of physical force or power, threatened or actual, against oneself, another person, or against a group or community, that either results in or has a high likelihood of resulting in injury, death, psychological harm, maldevelopment or deprivation.

From this definition, there are two requirements for something to be "violence", and of these must have been met for a post to be coded as violent.

1. It must have a threat or actual use of physical force or power. Since this work looks at social media, our work primarily focused on threats.

2. The threat must have a high likelihood of resulting in some type of harm.

These two factors help narrow the criteria for a post being labelled as violent. The WHO report, from which the definition originates, looks broadly at violence, including from nation-states against groups. We are focusing on social media posts made by individuals. Thus, while a person may make a call for nation-state violence (e.g. "The United States should invade Canada and kill everyone"), this is not an actual threat of violence because a regular social media user does not have any power to deploy the US military nor control over their actions. Note that if President Trump had made this post, for example, it could be considered violence because he had the power to do such a thing.

While the definition requires a high likelihood of harm, this does not mean that the threatener has to realistically be able to carry out their threat. If someone tweets "I am going to kill you", that would be coded as violence, even if they live around the world and could never get to the person they are targeting. The death threat, even if unlikely to be carried out, is a threat of physical force that is likely to result in psychological harm.

We would not classify a post as violent if it is reporting on an act, e.g. "A man killed 5 people today with an AK47 in a shooting at a mosque.". While the act described is violent, the post is not endorsing or supporting that violence. Note that this is a difference between our dataset and the YouTube dataset.

Offensive posts and deeply disturbing content are not necessarily violent. For example, one post in the corpus reads "I suggest that we start acting like n****rs and just take what we want. I stole this wrist band from a n****rs vehicle just now. [redacted by paper authors]" Despite the offensive content and language, this is not a threat (or use) of physical force of power. While seeing this post may cause psychological harm to certain groups, it would not qualify as violence because it is missing the threat/use of force/power component.

If a post is considered violent, we additional labelled whether the post was a threat of sexual or political violence. Sexual violence includes threats of sexual assault or physical violence taking place in a sexual context. Political violence describes violence against an opposing political group that targets an individual because of their politics, or that is designed to achieve a political goal.

For example, a post that read "Hang the traitors" would be labeled political violence. "Kill every liberal" would also.

There are some posts that, if read without context, may seem violent but, in context, are not. These include things like:

- Satire: posts that subtly mock violence or other groups who might employ it

- Humor: posts that are making jokes that may celebrate violence but that do not mean it seriously

Detecting these nuances takes careful human judgment. Our team trained on this, but we acknowledge these types of posts will be difficult to distinguish from actually violent posts for machine learning algorithms.

Posts were pulled from three platforms. Incel.is is the largest forum for the incel (involuntary celibate) community. On the surface, this online subculture is for men who struggle to form romantic relationships with women. In reality, incel forums are hotbeds for misogyny, racism, homophobia, self-hatred, violent fantasies, and sexual entitlement. A number of self-described incels have committed mass murder, and their forums have hosted videos and livestreams of mass shootings, and encouragement for more violence. They have been banned from popular forum hosts like reddit for repeated violations of the terms of service. We selected our pool of posts from the incels.is forum data available from [8].

The reddit politosphere is a dataset that covers over 600 subreddits (i.e. online forums) related to politics over 12 years from 2008 through 2019 [10]. This intentionally excludes Covid and 2020 US election-related posts.

Parler was a popular right wing microblogging platform, similar to Twitter. It was especially popular in 2020 and was a prominent source of information and planning around the January 6, 2021 insurrection. We randomly selected posts from the Parler dataset available from [3].

Coders trained as a group on sample data through several iterations until we achieved consensus on how to interpret the codebook. Each post was labeled by three coders, allowing a majority rule vote to determine the label. Because we had three coders, we calculated the Fleiss Kappa to measure inter rater reliability. On this data, k=0.48, indicating moderate to substantial agreement [14].

The Fleiss Kappa of 0.48 indicates moderate agreement among annotators. Out of the 30,000 total posts coded, 676 (approximately 2.3%) had some form of disagreement among the three annotators, necessitating resolution through majority voting. These disagreements typically fell into several categories: (1) differences in interpreting implied versus explicit threats, (2) varying assessments of the credibility or likelihood of harm from specific threats, (3) challenges in distinguishing between offensive but non-violent content and actual threats, and (4) disagreements about satirical or humorous content that references violence.

For instance, annotators occasionally disagreed on whether posts containing hypothetical scenarios (e.g., "If X happens, then people should Y...") constituted actual threats of violence. Similarly, determining whether comments that employed violent metaphors or hyperbole (e.g., "destroying" an opponent in debate) met the threshold for actual threats of physical force presented challenges. The majority voting approach helped ensure that borderline cases were resolved consistently, with only posts receiving at least two "violent" annotations being classified as violent in the final dataset.

Overwhelmingly, the posts in this dataset are not violent. Of 30,000 posts, 243 were labeled as violent by the majority of coders. This is to be expected because even on platforms without moderation where violent language is common, it should be expected that most posts are conversational and not explicit threats. For comparison, the YouTube Threat Corpus [10] found a higher but still relatively small proportion of violent content, with 4.8% of sentences containing violent threats. While this small sample presents certain limitations for training complex models, it provides a high-precision set of examples that can serve as reliable positive instances for classification tasks.

There were differences in violence levels between platforms. The incels.is data had violence 73 of 9,936 posts (0.73%). Of these, 11 (15.1%) were sexually violent and only 2 (2.74%) were politically violent. On Parler, 158 of 10,093 (1.56%) of posts were violent, with 102 (64.6%) containing political violence and only one (0.63%) containing sexual violence. Finally, on reddit - a moderated platform that does not allow violence - only 12 of 9,973 (0.12%) posts were violent and 100% of these were politically violent, with one of these also containing sexual violence.

The dataset is available for download on the Harvard Dataverse at doi.org/10.7910/DVN/ANGOX0

## 4 Training and Cross-Platform Performance

Our InViS dataset contains violence data from three platforms. Because there are differences among the platforms we studied, there are questions about how well this dataset can support machine learning that works across platforms. Perhaps there are substantial differences that make the data too diverse to pick up signals across platforms. Similarly, while our coding scheme varied from that used in the YouTube dataset, they capture similar concepts. There is a parallel question about whether models trained on InViS is useful for detecting violence in the YouTube dataset and vice versa. This section performs a cross-platform and cross-dataset analysis.

As noted above, the vast majority of posts in both datasets are not violence. Such a dramatic class imbalance inhibits a meaningful analysis of machine learning results. To address this, we created balanced data subsets with all the violent posts and an equal number of randomly selected non-violent posts. This approach allows us to more accurately assess how much signal is present in our data. The InViS dataset had 484 total posts and the YouTube dataset had 2,774.

We trained models using the Multinomial naive bayes for text data in Weka. We conducted a 10-fold cross validation to assess performance. On the InVis dataset, we achieved 73.6% accuracy with a precision of 0.736, recall of 0.736, and F1 score of 0.735. The ROC AUC was 0.807. On the YouTube dataset, we achieved an accuracy of 82.2% with precision of 0.824, recall of 0.823, F1 score of 0.822, and a ROC AUC of 0.898. Both results show good performance within datasets, including within the InViS dataset that includes posts from multiple platforms.

For our machine learning analysis, we used a bag-of-words approach with Weka's Multinomial Naive Bayes implementation for text classification. We applied default parameters, including word tokenization with standard delimiters, no stemming, no stopwords removal, and a minimum word frequency of 3. This approach was selected for its effectiveness with text classification tasks and to establish a baseline for cross-platform transferability. It's important to note that the performance metrics reported should be interpreted with appropriate caution. The 95% confidence intervals for our accuracy measurements are approximately ±4.0% for the InViS dataset experiments and ±1.9% for the YouTube dataset experiments, reflecting the relative sizes of these balanced datasets. These intervals indicate that while the cross-platform performance is encouraging, there remains some uncertainty, particularly for the smaller InViS dataset with its limited number of positive examples.

We next conducted a cross-dataset analysis, training on one dataset and using the other as a test set. In both cases, we used the balanced subsets for training and testing. For models trained on the InViS data and tested on the YouTube subset, we achieved 72.3% accuracy with precision of 0.726, recall of 0.723, F1 score of 0.722, and a ROC AUC of 0.795. For models trained on the YouTube data and tested on the InViS data, we achieved an accuracy of 68.0%, precision of 0.720, recall of 0.680, F1 score of 0.665, and ROC AUC of 0.736. While performing slightly worse than the within-dataset tests, both results show good performance training in one domain and testing in another.

Finally, we combined the balanced YouTube and InViS subsets into a single dataset and performed a 10-fold cross validation. This achieved 81.0% accuracy with a precision of 0.811, recall of 0.810, an F1 score of 0.810, and ROC AUC of 0.883.

## 5 Discussion and Conclusions

Our analysis suggests potential for cross-platform applicability of violence detection models, though with important limitations that must be acknowledged. The moderate performance of models trained on the InViS dataset when tested on YouTube data (72.3% accuracy) and vice versa (68.0% accuracy) indicates that there may be some shared linguistic patterns in violent content across platforms, despite variations in platform cultures and moderation policies. However, these results should not be overstated—the performance degradation in cross-platform testing compared to within-platform evaluation signals important platform-specific differences that more sophisticated models would need to address.

The small number of violent posts identified in our dataset (243 out of 30,000, or 0.81%) presents a significant limitation for training complex models, particularly deep learning approaches that typically require larger volumes of training data. This scarcity of positive examples, while reflecting the real-world rarity of explicitly violent content, constrains the generalizability of our findings and would likely necessitate data augmentation or transfer learning techniques for practical applications.

Qualitative analysis of misclassified examples reveals certain types of violent content that proved particularly challenging to detect across platforms. Indirect threats that use euphemisms or coded language were frequently missed by our models, as were culturally-specific references to violence that might be obvious to human moderators but opaque to algorithmic detection. Political violence expressed through historical references (e.g., guillotines, certain historical revolutions) presented particular challenges, as did threats embedded within longer, otherwise non-violent posts.

These results challenge an implicit assumption in content moderation: that platform-specific training data is necessary for effective content detection. While platform-specific models might achieve marginally better performance, our findings suggest that the fundamental linguistic markers of violent content are sufficiently universal to enable effective cross-platform detection. This has practical implications for platforms developing content moderation systems, particularly smaller platforms that may lack extensive labeled training data. Rather than building platform-specific datasets from scratch, platforms might leverage existing violence detection datasets, potentially supplementing them with smaller platform-specific training sets.

Our work makes several key methodological contributions to the quantitative description of online violence. The balanced dataset approach we employed addresses a fundamental challenge in violence detection research: the extreme rarity of violent content in natural social media data (with rates below 2% even on minimally moderated platforms). While this balancing was necessary for meaningful model evaluation, it's important to note that it does not reflect natural content distribution - rather, it provides a clearer lens for examining the detectability of violent content when it does occur. The moderate to substantial inter-rater reliability ($k=0.48$) we achieved suggests that our WHO-based coding framework provides a workable foundation for consistent content classification across different platforms and contexts.

The development and deployment of automated violence detection systems raises several important ethical considerations. False positives in violence detection can lead to unjustified content removal and potential chilling effects on free expression, particularly for marginalized communities whose communication styles may be unfairly flagged. Conversely, false negatives can leave harmful content unaddressed, potentially contributing to real-world harm or the normalization of violent rhetoric. Our annotation process itself raised ethical challenges, as annotators were exposed to disturbing content, including explicit threats of violence against vulnerable groups. We implemented safeguards to mitigate potential harm to annotators, including clear guidelines for taking breaks and the right to skip particularly disturbing content. The dataset itself presents considerations regarding representation, as different communities, languages, and cultural contexts may express and interpret violence in ways not captured by our sample. Finally, we acknowledge that violence detection tools, while intended to reduce harm, can be misused for censorship or surveillance if deployed without appropriate oversight and safeguards.

Our multi-platform sampling strategy offers a novel approach to dataset construction in content analysis. Instead of attempting to draw causal conclusions about why different platforms exhibit different patterns of violent content, we focused on documenting and quantifying these variations in detail. This approach revealed clear descriptive differences in both the prevalence and nature of violent content across platforms, while simultaneously demonstrating the feasibility of cross-platform detection. The inclusion of both mainstream and fringe platforms in our sample provides a broader view of how violent content manifests across different corners of the social media landscape.

**ACKNOWLEDGMENTS**

This research was supported by the National Science Foundation (Award #2331257) and a University of Maryland Catalyst Fund New Directions Award.